\documentclass[akbc,twoside,11pt,lettersize]{article}
\usepackage{akbc}
\usepackage{color}
\usepackage{xcolor}
\usepackage{booktabs}
\usepackage{multirow}
\usepackage{graphicx}
\usepackage{pgfplots}
\usepackage{paralist}
\usepackage{tabularx}
\usepackage{makecell}
\usepackage{stackengine}
\usepackage{rotating}
\usepackage[inline]{enumitem}
\usepackage{caption}
\usepackage{subcaption}
\usepackage{soul}
\usepackage{wrapfig}
\usepackage{lipsum}

\akbcheading
\ShortHeadings{How Context Affects Language Models' Factual Predictions}{F. Petroni, P. Lewis, A. Piktus, T. Rockt\"aschel, Y. Wu, A. H. Miller, S. Riedel}

\definecolor{myorange}{RGB}{244, 125, 35}
\definecolor{mygreen}{RGB}{0, 140, 72}
\definecolor{myblue}{RGB}{24, 90, 169}
\definecolor{myred}{RGB}{238, 46, 47}

\newcommand{\rel}[1]{\verb~#1~}

\newcommand{\mask}{\textsc{[Mask]}}
\newcommand{\sep}{\textsc{[SEP]}}
\newcommand{\eos}{\textsc{eos}}
\newcommand*{\red}{\textcolor{red}}

\newcommand{\lama}{\textsc{LAMA}}
\newcommand{\bert}{\textsc{B}}
\newcommand{\drqa}{\textsc{DrQA}}
\newcommand{\ret}{\textsc{B-ret}}
\newcommand{\gen}{\textsc{B-gen}}
\newcommand{\ora}{\textsc{B-ora}}
\newcommand{\adv}{\textsc{B-adv}}

\newcommand{\robertatext}{\textsc{RoBERTa}}
\newcommand{\berttext}{\textsc{BERT}}

\newcommand{\question}{\ensuremath{q}}
\newcommand{\context}{\ensuremath{c}}
\newcommand{\answer}{\ensuremath{a}}

\newcommand{\eg}{\textit{e.g.}}
\newcommand{\ie}{\textit{i.e.}}
\newcommand{\patone}{\ensuremath{P@1}}

\newcommand{\nsp}{NSP}

\finalcopy % Uncomment for camera-ready version, but NOT for submission.
\begin{document}

\title{How Context Affects Language Models' Factual Predictions}

\newcommand{\fair}{$^1$}
\newcommand{\ucl}{$^2$}
\newcommand{\uclfair}{$^{1,2}$}

\author{\name Fabio Petroni\fair{} \email fabiopetroni@fb.com \\
        \name Patrick Lewis\uclfair \email plewis@fb.com \\
        \name Aleksandra Piktus\fair \email piktus@fb.com \\
        \name Tim Rockt\"aschel\uclfair \email rockt@fb.com \\
        \name Yuxiang Wu\ucl \email yuxiang.wu.18@ucl.ac.uk \\
        \name Alexander H. Miller\fair \email  ahm@fb.com \\
        \name Sebastian Riedel\uclfair \email sriedel@fb.com \\
       \addr \fair{}Facebook AI Research \\
       \addr \ucl{}University College London \\ }

\maketitle

\begin{abstract}
When pre-trained on large unsupervised textual corpora, language models are able to store and retrieve factual knowledge to some extent, making it possible to use them directly for zero-shot cloze-style question answering.
However, storing factual knowledge in a fixed number of weights of a language model clearly has limitations.
Previous approaches have successfully provided access to information outside the model weights using supervised architectures that combine an information retrieval system with a machine reading component.
In this paper, we go a step further and integrate information from a retrieval system with a pre-trained language model in a purely unsupervised way.
We report that augmenting pre-trained language models in this way dramatically improves performance and that the resulting system, despite being unsupervised, is competitive with a supervised machine reading baseline.
Furthermore, processing query and context with different segment tokens allows BERT to utilize its Next Sentence Prediction pre-trained classifier to determine whether the context is relevant or not, substantially improving BERT's zero-shot cloze-style question-answering performance and making its predictions robust to noisy contexts.
\end{abstract}

\section{Introduction}

Pre-trained language models such as BERT~\cite{devlin2018bert} and RoBERTa~\cite{liu2019roberta} enabled state-of-the-art in many downstream NLP tasks~\cite{wang-etal-2018-glue,wang2019superglue,wu2019zero}.
These models are trained in an unsupervised way from large textual collection and recent work~\cite{petroni2019language,DBLP:journals/corr/abs-1911-12543,DBLP:journals/corr/abs-1912-13283,devlin2018bert} has demonstrated that such language models can store factual knowledge to some extent.
However, considering the millions of documents and facts in Wikipedia\footnote{\url{https://en.wikipedia.org/wiki/Wikipedia:Statistics}} and other textual resources, it unlikely that a language model with a fixed number of parameters is able to reliably store and retrieve factual knowledge with sufficient precision~\cite{realm}.

One way to get around this is to combine machine reading with an information retrieval (IR) system~\cite{chen2017reading,realm}. Given a question, the IR system retrieves relevant contexts which are subsequently processed by a reading component.
In the case of DrQA~\cite{chen2017reading}, the retriever is fixed and the reading component is trained based on retrieved contexts, whereas in REALM~\cite{realm} the IR component is trained alongside the reader during both pre-training and subsequent fine-tuning.

In this paper, we go a step further and forego supervised fine-tuning. Instead, we consider the purely unsupervised case of augmenting a language model with retrieved contexts at test time.
We demonstrate that augmenting pre-trained language models with such retrieved contexts dramatically improves unsupervised cloze-style question answering, reaching performance that is on par with the supervised DrQA approach.
In addition to being unsupervised, using a pre-trained language model like BERT instead of a trained machine reading component has several other advantages. Since BERT is not an extractive QA model, it is able to utilize contexts that contain relevant information but do not contain the answer span directly.
More importantly, we find that via the next-sentence prediction objective BERT is able to ignore noisy or irrelevant contexts.

In summary, we present the following core findings:
\begin{inparaenum}[i)]
\item augmenting queries with relevant contexts dramatically improves BERT and RoBERTa performance on the \lama{} probe \cite{petroni2019language}, demonstrating unsupervised machine reading capabilities of pre-trained language models;
\item fetching contexts using an off-the-shelf information retrieval system is sufficient for BERT to match the performance of a supervised open-domain QA baseline;
\item BERT's next-sentence prediction pre-training strategy is a highly effective unsupervised mechanism in dealing with noisy and irrelevant contexts.
\end{inparaenum}
The code and data to reproduce our analysis will be made publicly available.
\section{Related Work}
\label{sec:rw}

\paragraph{Language Models and Probes} 
With the recent success of pre-trained language models like BERT~\citep{devlin2018bert} and its variants~\citep{liu2019roberta,DBLP:conf/acl/SeoLKPFH19,DBLP:journals/corr/abs-1910-10683,lewis2019bart}, it becomes increasingly important to understand what these models learn. A variety of ``probes'' have been developed to analyse the syntactic structures, such as syntax trees~\citep{DBLP:conf/emnlp/MarvinL18,DBLP:conf/naacl/HewittM19, DBLP:journals/corr/abs-1906-04284}, negative polarity items~\citep{warstadt2019linguistic, DBLP:conf/emnlp/WarstadtCGPBNAB19}, semantic fragments~\citep{DBLP:journals/corr/abs-1909-07521}, function words~\citep{DBLP:conf/starsem/KimPPXWMTRLDBP19}, and many other linguistic phenomena~\citep{DBLP:conf/acl/TenneyDP19,DBLP:conf/iclr/TenneyXCWPMKDBD19,de2020decisions}.
To measure the factual knowledge present in these pre-trained language models, \citet{petroni2019language} propose the \lama{} benchmark which tests the models with cloze-style questions constructed from knowledge triples.
\citet{DBLP:journals/corr/abs-1911-12543} later extends \lama{} by automatically discovering better prompts, \citet{kassner2019negated} add negated statements, \citet{poerner2019bert} filter out easy-to-guess queries, and \citet{DBLP:journals/corr/abs-1912-13337, DBLP:journals/corr/abs-1912-13283, DBLP:journals/corr/abs-1911-11641} develop further probes for textual reasoning. 
Pre-trained language models have also been fine-tuned and adapted to be used as information retrieval systems~\citep{DBLP:journals/corr/abs-1903-10972,DBLP:conf/emnlp/YilmazWYZL19,DBLP:conf/acl/SeoLKPFH19}.

\paragraph{Open-Domain QA}
Open-domain QA aims at answering questions without explicitly knowing which documents contain relevant information. 
Open-domain QA models often involve a retriever to find relevant documents given a question, and a reader to produce the answers~\citep{chen2017reading}. Works in this areas mostly focus on enhancing retrieval quality~\citep{choi-etal-2017-coarse,DBLP:conf/aaai/WangYGWKZCTZJ18,DBLP:conf/iclr/WangY0ZGCWKTC18,DBLP:conf/acl/SunLLJ18,DBLP:conf/acl/SocherZXM18,DBLP:conf/emnlp/LeeYKKK18,lee-etal-2019-latent,DBLP:conf/iclr/DasDZM19,DBLP:journals/corr/abs-1909-07597}, improving answer aggregation~\citep{clark2017simple,DBLP:conf/iclr/WangY0ZGCWKTC18,DBLP:conf/emnlp/LeeYKKK18,DBLP:conf/aaai/PangLGXSC19}, and accelerating the whole pipeline~\citep{DBLP:conf/acl/SeoLKPFH19}. Recently, \citet{realm} show that augmenting language model pre-training with a knowledge retriever induces performance gains on open-domain QA tasks. Our work differs from previous works in open-domain QA in two ways: \begin{inparaenum}[i)] 
\item we consider a fully unsupervised setting using a pre-trained language model and an off-the-shelve information retrieval system, 
\item our aim is to assess the prediction of factual knowledge in this setup rather than to improve open-domain question answering in general.
\end{inparaenum}

\section{Methodology}

Given a cloze-style question \question{} with an answer \answer{}, we assess how the predictions from a language model change when we augment the input with contexts \context{}.
In this section, we describe the datasets we use to source $(\question, \answer)$ pairs, as well as various methods of generating context documents $\context$.

\subsection{Datasets}

\begin{wraptable}{r}{0.5\textwidth} 
%\begin{table}[t!]
            \centering
            %\resizebox{\textwidth}{!}{   
            \begin{tabular}{llcc}
                \toprule
                \multirow{2}{*}{Corpus} & \multirow{2}{*}{Relation} & \multicolumn{2}{c}{Statistics}  \\ 
                     & & \#Facts & \#Rel  \\ 
         \midrule
\multirow{4}{*}{Google-RE}  & \rel{birth-place} & 2937 & 1  \\ 
&\rel{birth-date} & 1825 & 1 \\ 
&\rel{death-place} & 765 & 1  \\ 
\cmidrule{2-4} 
& Total & 5527 & 3  \\
\midrule 
\multirow{4}{*}{T-REx}  & $1$-$1$ & 937 & 2  \\
&$N$-$1$ & 20006 & 23  \\
& $N$-$M$ & 13096 & 16  \\
\cmidrule{2-4} 
& Total & 34039 & 41  \\
\midrule 
SQuAD & Total & 305 & -  \\
\bottomrule 
\end{tabular}
%}
\caption{ Statistics for the LAMA data.}
\label{tab:LAMAstat}
%\end{table}
\end{wraptable}

We use the \lama{}\footnote{\url{https://github.com/facebookresearch/LAMA}} probe in our experiments~\cite{petroni2019language}, a collection of cloze-style questions about real world relational facts with a single token answer. Each question is accompanied by snippets of text from Wikipedia that are likely to express the corresponding fact.
Although there are several cloze-style QA datasets (some listed in Section \ref{sec:rw}) we decided to use \lama{} because: (1) the nature of the \lama{} data is aligned with the relational knowledge focus or our analysis (\ie, given a subject and a relation predict the object) and (2) each data point is aligned by construction with relevent contextual information.
We consider the Google-RE\footnote{\url{https://code.google.com/archive/p/
relation-extraction-corpus}} (3 relations, 5527 facts), T-REx \cite{elsahar2019t} (41 relations, 34039 facts) and SQuAD~\cite{rajpurkar2016squad} (305 questions manually translated in cloze-style format) subsets of the probe. More detailed statistics for the \lama{} data considered are reported in Table~\ref{tab:LAMAstat}.
For the \textsc{RoBERTa} results, we trim the \lama{} dataset (by about 15\%) such that all answers are in the model's vocabulary, so \textsc{BERT} and \textsc{RoBERTa} numbers in this paper should not be directly compared as they consider slightly different subsets of the data.

\subsection{Baselines}
We consider \drqa{}~\cite{chen2017reading}, a popular system for open-domain question answering. The overall pipeline consists of two phases: (1) a TF-IDF document retrieval step, where the model finds relevant paragraphs from Wikipedia and (2) a machine comprehension step to extract the answer from those paragraphs. The machine comprehension component is trained with supervision on SQuAD v1.1~\cite{rajpurkar2016squad}. In order to apply \drqa{} to the \lama{} probe, we take inspiration from \cite{levy2017zero} and map each cloze-style template to a natural question template (\eg, ``\textsc{X} was born in \mask{}'' to ``Where was \textsc{X} born?''). We constrain the predictions of \drqa{} to single-token answers as in \citet{petroni2019language}.
Our results for \drqa{} and \textsc{BERT} are directly comparable with the other baselines in \citet{petroni2019language}.

\subsection{Language Models}
Among the constellation of language models that have been proposed in recent years we consider \textsc{BERT}~\cite{devlin2018bert} since it is one of the post popular and widely used at the time of writing.\footnote{\url{https://huggingface.co/models}}
Moreover, the large cased version of the \textsc{BERT} model is the best performing LM on the \lama{} probe among those considered in \citet{petroni2019language}.
We additionally consider the large version of the \textsc{RoBERTa} model~\cite{liu2019roberta}.
Both \textsc{BERT} and \textsc{RoBERTa} have been trained on corpora that include Wikipedia. While \textsc{BERT} uses two pre-training strategies, Mask Language Modelling (MLM) and Next Sentence Prediction (NSP), \textsc{RoBERTa} considers only the MLM task.
We produce a probability distribution over the unified vocabulary of \citet{petroni2019language} for the masked token in each cloze-style questions and report the average precision at $1$ (P@1).

\subsection{Contexts}

We enrich cloze statements with different types of contextual information.
We explicitly distinguish cloze question \question{} and context \context{} in the input according to the model. For \textsc{BERT}, we use different segment embeddings, index 0 for \question{} and 1 for \context{}, and insert the separator token (\ie, \sep) in between. For \textsc{RoBERTa}, which is not equipped with segment embeddings, we use the end of sentence (\eos) token to separate \question{} and \context{}. We addidionally performed some experiments without this clear separation of query and context, but considering them as concatenated in a single segment (or wihtout the \eos{} token in between).
The input is truncated to $512$ tokens.

\subsubsection{Oracle Contexts}

We provide an oracle-based (\textsc{ora}) context in order to assess the capability of LMs to exploit context that we know is relevant to the entity in the question. Concretely, we use the Wikipedia snippet accompanying each example in the \lama{} probe, truncated to at most five sentences. This context often contains the true answer and always contains related true information.

\subsubsection{Sourcing Relevant Contexts}

Relevant context is often not available and must be automatically sourced by the model~\cite{chen2017reading,clark2017simple}. In this scenario, we consider two possible approaches: using an information retrieval engine (\textsc{ret}) or generating the context with an autoregressive LM (\textsc{gen}) \cite{radford2019language}.
For the retrieval case, we use the first paragraph from \drqa{}'s retrieval system as context. 
For the generative case, taking inspiration from the study of \citet{massarelli2019decoding}, we consider a 1.4B parameters autoregressive language model trained on \textsc{CC-NEWS}~\cite{liu2019roberta}. This model has been shown to generate more factual text with respect to others trained on different corpora, including Wikipedia. For each question in \lama{}, we use the natural question template as prefix to condition the generation, and generate five sentences using the delayed beam search strategy~\cite{massarelli2019decoding}. These results may be quite related to the entity in the query, though they may not always be completely factual.

\begin{table*}[t!]
            \centering
            %\resizebox{\textwidth}{!}{   
            \begin{tabular}{llcccccc}
                \toprule
                \multirow{2}{*}{\lama} & \multirow{2}{*}{Relation} &  & & \multicolumn{3}{|c|}{\textit{open domain sourced context}}   \\ 
                   &  & \bert & \adv & \multicolumn{1}{|c}{\gen} & \drqa & \multicolumn{1}{c|}{\ret} & \ora  \\ 
         \midrule
\multirow{4}{*}{Google-RE}  & \rel{birth-place} & 16.1 & 14.5 & 8.5 & \textbf{48.6} & 43.5 & \textit{70.6}   \\ 
&\rel{birth-date} & 1.4 & 1.4 & 1.4 & 42.9  & \textbf{43.1} & \textit{98.1}   \\ 
&\rel{death-place} & 14.0 & 12.6 & 6.0 & \textbf{38.4}  & 35.8 & \textit{65.1} \\ 
\cmidrule{2-8} 
& Total  & 10.5 & 9.5 & 5.3 & \textbf{43.3}  & 40.8 & \textit{78.0}  \\
\midrule 
\multirow{4}{*}{T-REx} & $1$-$1$ & 74.5 & 74.5 & 71.3 & 55.2  &\textbf{81.2}  & \textit{91.1} \\
&$N$-$1$ & 34.2 & 33.8 & 32.7 & 30.4  & \textbf{47.5}  & \textit{67.3} \\
& $N$-$M$ & 24.3 & 23.6 & 23.8 & 15.4 & \textbf{32.0} & \textit{52.4}\\
\cmidrule{2-8} 
& Total  & 32.3 & 31.8 & 31.1 & 25.8 & \textbf{43.1} & \textit{62.6} \\
\midrule 
% ConceptNet & Total & 11458 & 16 & 4.8 & - & - & - &  19.2 & 16.3 & 16.6 & - \\
% \midrule 
SQuAD &  & 17.4 & 17.4 & 15.8 & \textbf{37.5}  & 34.3  & \textit{61.7}\\
\midrule
    \multicolumn{2}{c}{\textit{weighted average}} & 30.5 & 30.0 & 29.0 & 27.2  & \textbf{42.8}  & \textit{63.6}\\
\bottomrule 
\end{tabular}
%}
\caption{Mean precision at one (P@1) for the \drqa{} baseline, BERT-large on context-free cloze questions (\bert) and on adversarial (\adv), generated (\gen), retrieved (\ret) and oracle (\ora) context-enriched questions on the relational \lama{} probe. The fully unsupervised \ret{} is competitive with the supervised \drqa{} system and is dramatically better than the context-free baseline. We weight the average per number of relations (3 for Google-RE, 41 for T-REx and we consider SQuAD as a single contribution). Pairwise sign tests per relation show statistically significant differences (p-value below 1e-5) between: \ret{} and all other results; \ora{} and all other results.     }
\label{tab:mainresults}
\end{table*}

% ORACLE 
% 1 sentence Google-RE: 49.2, 91.4, 12.0 - 50.9
% 3 sentences Google-RE: 51.4, 96.5, 19.9 - 55.9

% 1 sentence T-REx: 89.5, 55.6, 51.6 - 55.7
% 3 sentences T-REx: 90.1, 57.6, 53.3 - 57.5

% GENERATED
% 5 sentences Google-RE DelayedBS: 6.0, 1.0, 1.7 - 2.9
% 5 sentences Google-RE BS: 10.2, 1.0, 2.0 - 4.4
% 5 sentences Google-RE BS white: 5.7, 1.0, 2.8 - 3.1

\subsubsection{Adversarial Contexts}
We provide an uninformative context in order to test the ability of the model to ignore irrelevant context that is not useful for answering the query. We do this by randomly sampling an oracle context from a different question that has the same relation type but a different answer $a'$.
This results in a context document that refers to a different subject entity but contains a distracting and semantically plausible answer $a'$. Table \ref{tab:trexexamplesbert} shows some examples of adversarial contexts.

\section{Results}
\label{sec:results}

\if 0
\begin{figure}[t!]
    \centering
    \includegraphics[width=0.6\linewidth]{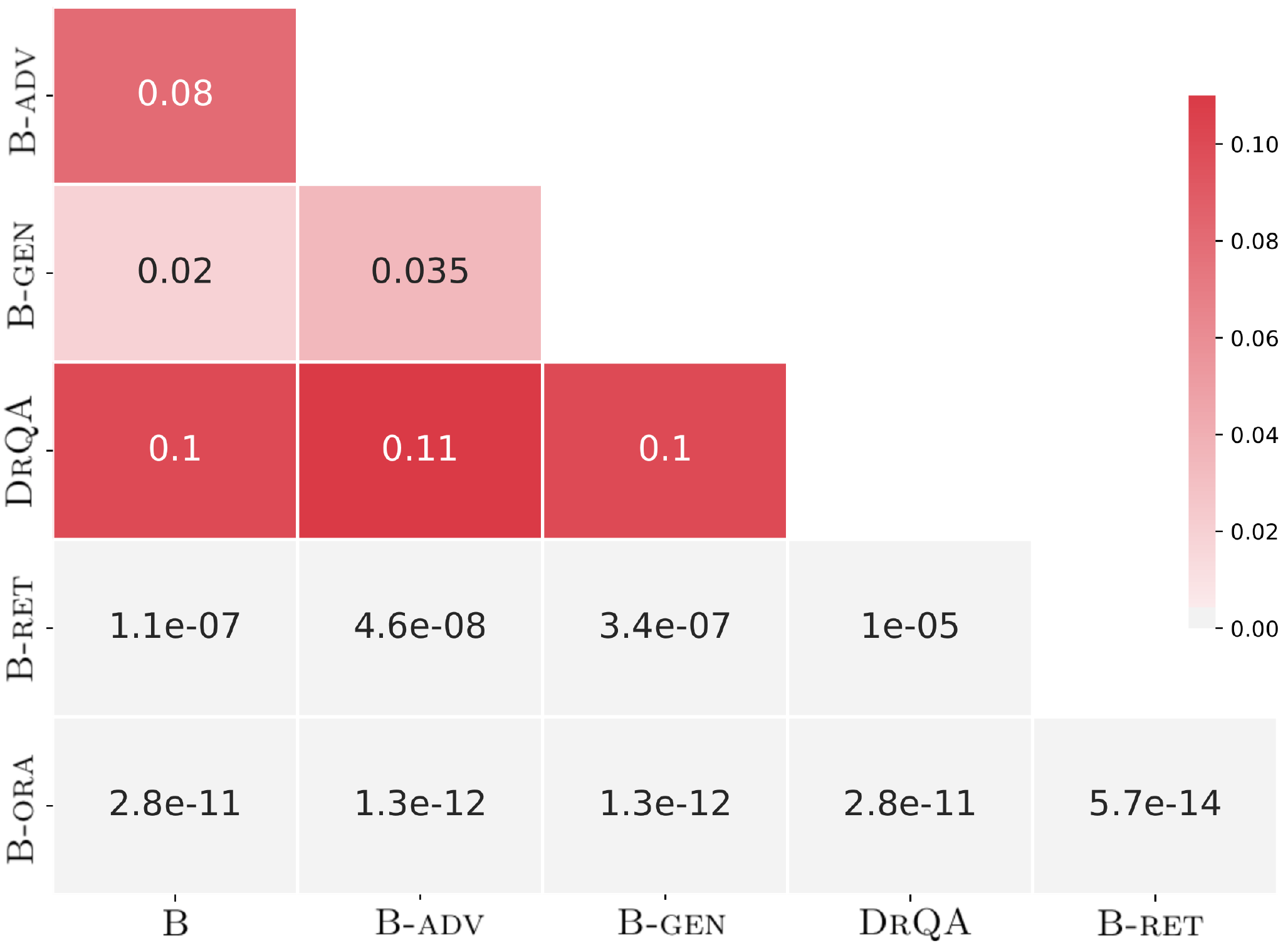}
    \caption{\red{Pairwise statistical significance for the results presented in Table \ref{tab:mainresults}, using the sign test across relations. Each cell reports the p-value of the corresponding pair. The improvements achieved by \ret{} and \ora{} are statistically significant (p-value smaller than the alpha level of 0.05).}}
    \label{fig:statistical significance}
\end{figure}
\fi

\begin{figure}
\centering
    \begin{subfigure}{0.45\textwidth}
    \centering
        \begin{tikzpicture}[scale=0.79]
        \begin{axis}[
            xlabel={$k$: number of retrieved paragraphs},
            ylabel={recall accuracy},
            xmin=1, xmax=10,
            ymin=0, ymax=70,
            xtick={1,2,3,4,5,6,7,8,9,10},
            ytick={0,10,20,30,40,50,60,70,80,90,100},
            legend style={at={(1,0.3)},anchor=south east},
            ymajorgrids=true,
            grid style=dashed,
        ]
         
        \addplot[
            color=myblue,
            mark=triangle*,
            mark size=2pt,
            line width=0.25mm
            ]
            coordinates {
            (1,45.28)
            (2,55.17)
            (3,58.99)
            (4,60.87)
            (5,61.85)
            (6,62.5)
            (7,62.92)
            (8,63.39)
            (9,63.68)
            (10,63.97)
            };
            \addlegendentry{Google-RE}

        \addplot[
            color=mygreen,
            mark=square,
            mark size=2pt,
            line width=0.25mm
            ]
            coordinates {
            (1,25.24)
            (2,32.8)
            (3,36.98)
            (4,39.9)
            (5,42.29)
            (6,44.08)
            (7,45.55)
            (8,46.63)
            (9,47.79)
            (10,48.7)
            };
            \addlegendentry{T-REx}

%        \addplot[
%            color=myblue,
%            mark=x,
%            mark size=1pt
%            ]
%            coordinates {
%            (1, 2.1)
%            (2, 3.49)
%            (3, 4.5)
%            (4, 5.41)
%            (5, 6.16)
%            (6, 6.84)
%            (7, 7.51)
%            (8, 8.24)
%            (9, 8.86)
%            (10, 9.46)
%            };
%            \addlegendentry{ConceptNet}
            
        \addplot[
            color=myred,
            mark=o,
            mark size=2pt,
            line width=0.25mm
            ]
            coordinates {
            (1,34.1)
            (2,41.97)
            (3,45.9)
            (4,47.87)
            (5,50.16)
            (6,51.15)
            (7,52.79)
            (8,53.44)
            (9,54.1)
            (10,55.08)
            };
            \addlegendentry{SQuAD}

        \end{axis}
        \end{tikzpicture}
    \caption{Percentage of times the answer appears in the top-$k$ retrieved paragraphs by \drqa. We use k=1 for our experiments as a single paragraph can already contain a large number of tokens.}
    \label{fig:retrecall}
\end{subfigure}
  \begin{subfigure}{.53\textwidth}
  %\begin{table}[t!]
%            \centering
            \resizebox{\textwidth}{!}{   
            \begin{tabular}{llcccc}
                \toprule
               \rotatebox[origin=c]{90}{P@1}    & \makecell{ \textit{answer}\\ \textit{in ctx} } & \multicolumn{1}{|c}{\adv} & \gen & \ret & \ora    \\ 
         \midrule
\multirow{3}{*}{\rotatebox[origin=c]{90}{\textit{better}}} & \textit{present} & 0.9 & 4.6 & 14.0  & 32.6 \\
& \textit{absent} & 2.4 & 2.5 & 3.2  & 1.4  \\
\cmidrule{2-6} 
& Total & 3.3 & 7.0 & 17.2  & 34.0  \\
\midrule
\multirow{3}{*}{\rotatebox[origin=c]{90}{\textit{worse}}}  &  \textit{present} & 0.6 & 2.0 & 2.4  & 3.5 \\
& \textit{absent}  & 3.1 & 6.2 &  3.9  & 0.1  \\
\cmidrule{2-6} 
& Total & 3.7 & 8.2 &  6.3  & 3.6  \\
\midrule
\multicolumn{2}{l}{\textit{\# better rel.}}  & 11 & 13 &  34  & 39  \\
\bottomrule %
\end{tabular}
}
%\caption{Origin of P@1 difference for T-REx as percentage of data points for which P@1 is \textit{better} (or \textit{worse}) than answering context-free questions, grouped by cases where context contains (\textit{yes}/\textit{no}) the answer.}
\caption{For T-REx, we report the percentage of time the model changes its output for the \textit{better} or \textit{worse} when the context is provided, grouped by the \textit{presence} or \textit{absence} of the answer in the provided context. \ret{} and \ora{} scored higher than the context-free model on most relations.}
\label{tab:origintrex}
%\end{table}

% (out of 41)

% RELATION P407 BETTER ZERO THAN ORACLE
% RELATION P1412 BETTER ZERO THAN ORACLE
    \end{subfigure}
\caption{}
\label{fig:double}
\end{figure}

The main results of our analysis are summarized in Table \ref{tab:mainresults}.
It shows the mean precision at one (P@1) for the \drqa{} baseline and BERT-large on the \lama{} probe enriched with different kinds of contextual information. 
Enriching cloze-style questions with relevant context dramatically improves the performance of BERT: \ora{} obtains $\times 7.4$ improvement on Google-RE, $\times 1.9$ on T-REx and  $\times 3.5$ on SQuAD with respect to using context-free questions (\bert{}). 
This clearly demonstrates BERT's ability to successfully exploit the provided context and act as a machine reader model.
Remarkably, no fine-tuning is required to trigger such behaviour.

When we rely on TF-IDF retrieved context (\ret), BERT still performs much better than in the no context setting. Overall, \ret{} results are comparable with \drqa{} on Google-RE and SQuAD and much higher on T-REx. 
This is particularly surprising given that \ret{}, unlike \drqa{}, did not receive any supervision for this task. Pairwise sign tests across relations show that the improvements for \ret{} and \ora{} are indeed statistically significant (p-value below 1e-5). 

Figure \ref{fig:retrecall} shows the recall of the IR system, which demonstrates that the answer is not present in many of the retrieved contexts, though often the context is still related to the same topic.
Table \ref{tab:origintrex} reports a detailed analysis of whether the answer is present in retrieved contexts and how that affects the model's predictions. 
We observe that most of the gain of \ret{} comes from cases in which the context contained the answer. However, there are also cases where the context does not explicitly mention the answer but BERT is still able to utilize the related context to help select the correct answer. Note that an extractive approach (such as \drqa{}) would have provided an incorrect answer (or no answer) for those cases.

\subsection{Adversarial Robustness}

\begin{figure}[t!]
    \centering
    \includegraphics[width=0.8\linewidth]{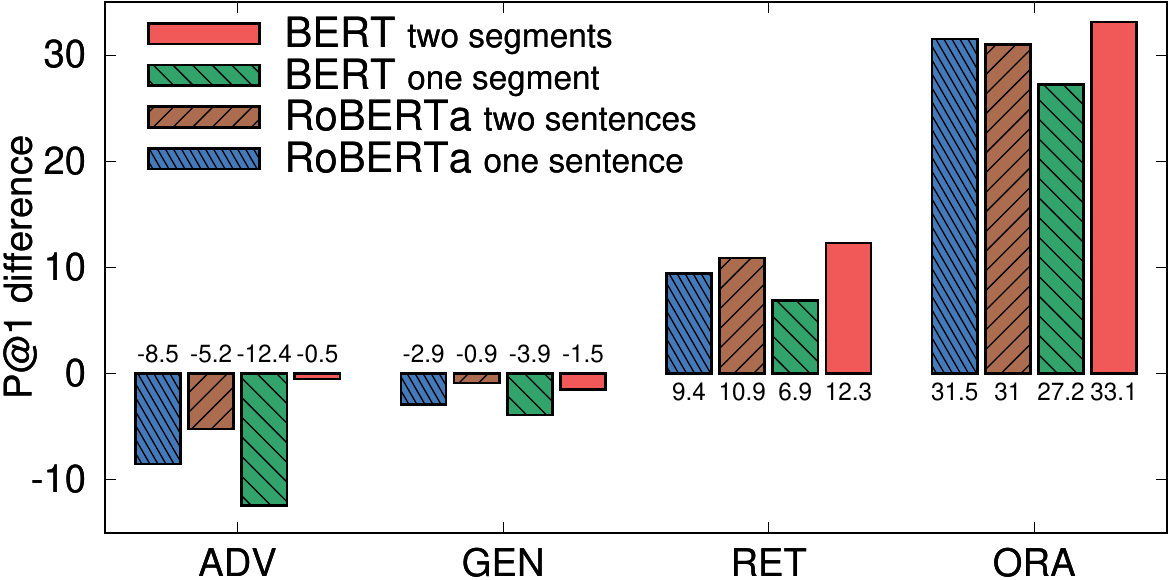}
    \caption{For each type of context considered, we report the change in P@1 relative to zero context, averaging results across relations. For each model we consider a concatenation of question and context as well as separating the two using separator tokens (\textsc{BERT}) or end of sentence tokens (\textsc{RoBERTa}). Separation dramatically improves both model's ability to ignore poor context and improves \textsc{BERT}'s performance in the presence of good context.}
    \label{fig:distributionp}
\end{figure}

The \adv{} column in Table \ref{tab:mainresults} shows the \lama{} \patone{} results for \berttext{} for adversarial contexts. 
\berttext{} is very robust, dropping only 0.5 \patone{} on average from the zero context baseline. 
However, as shown in Figure \ref{fig:distributionp}, this strong performance only occurs when the context and question are processed as two segments using \berttext{}'s separator tokens.
Using only one segment (that is, simply concatenating the input query and the context) leads to a severe drop of 12.4 \patone{} for \berttext{} (a 40.7\% relative drop in performance).
We also observe a consistent improvement in performance from one segment to two for retrieved and oracle contexts.

One possible reason for this phenomenon resides in the Next Sentence Prediction (\nsp{}) classifier of \berttext{}, learned with self-supervision during pretraining by training the model to distinguish contiguous (\ie, ``next sentence" pairs) from randomly sampled blocks of text. 
We hypothesize that the MLM task might be influenced by the \nsp{}'s output. Thus, \berttext{} might learn to not condition across segments for masked token prediction if the \nsp{} score is low, thereby implicitly detecting irrelevant and noisy contexts. 
A result that seems in line with this hypothesis is that \robertatext{}, which does not use \nsp{}, is more vulnerable to adversarial contexts and the difference between one and two sentences (for \robertatext{} separated by the \eos{} token) is much smaller.

\begin{table*}[t!]
    \centering
    \begin{tabular}{lccccc}
        \toprule
        %\lama &   \multicolumn{2}{|c|}{\textit{retrieved context}} &  &  \multicolumn{1}{|c}{} \\ 
        \scriptsize{(\% Next Sentence)} & \multicolumn{1}{c}{\adv} & \multicolumn{1}{c}{\gen} &  \multicolumn{1}{c}{\ret}  & \ora  \\ 
     \midrule
     Google-RE & 10.4 & 95.1 & 88.9  & 98.4  \\
     T-REx & 14.0 & 97.0 & 89.7 & 94.5  \\
     SQuAD & 11.9& 96.4  & 93.1  & 99.3  \\
    \bottomrule 
    \end{tabular}

\caption{Percentage of examples classified as `next sentences' according to BERT's NSP classifier for the different context types. The low number of `next sentence' classifications for \adv{} shows the model is able to recognize that adversarial contexts are unrelated and thus limit its influence on modeling the masked token in the query.}
\label{tab:table_nsp}
\end{table*}

\begin{figure}[!t]
    \centering
    \includegraphics[width=\linewidth]{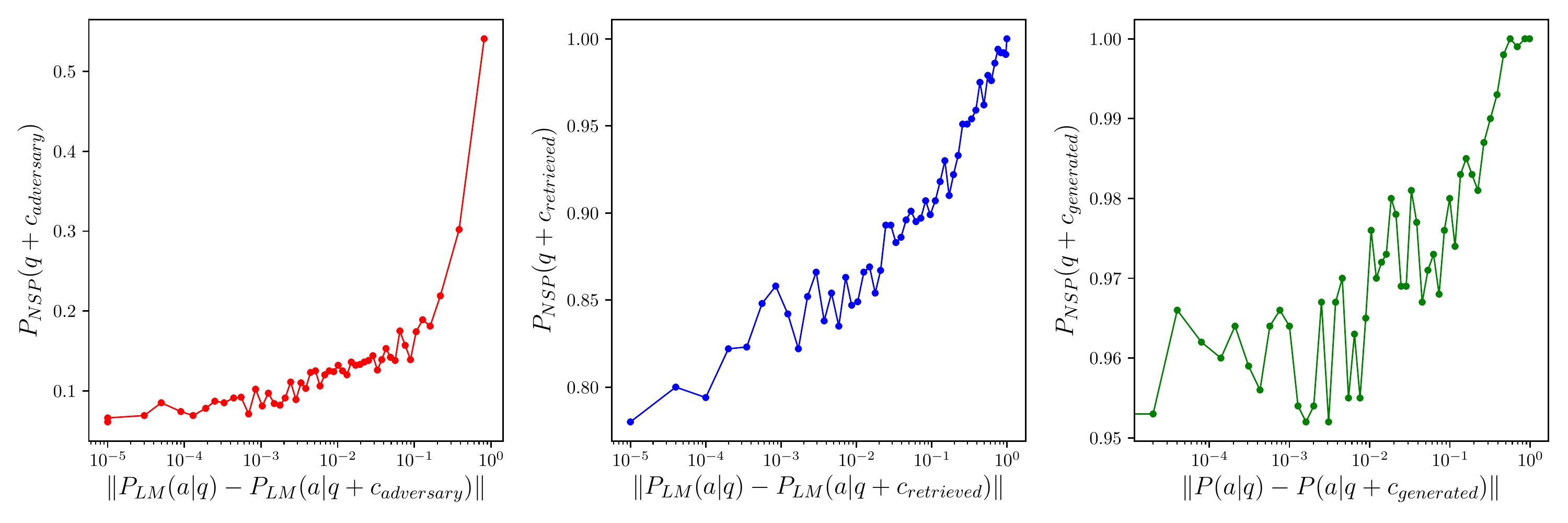}
    \caption{NSP Probabilities vs the change in LM probability upon appending contexts from the adversary (left), retrieval (mid), and generation (right) systems.
    At higher NSP probabilities, we see a higher larger increase to the probability mass placed on the correct answer in the presence of the context. That is, the more relevant that BERT thinks the context is, the more we see an increase to the likelihood of the true answer. This is exactly what we would want to see if we had hand-trained a relevance system ourselves, yet this instead emerges naturally from BERT's NSP pre-training loss.
    }
    \label{fig:nsp_correlation}
\end{figure}

To further investigate this hypothesis, we calculate the number of $(\context,\question)$ pairs classified by \berttext{} as ``next sentence" pairs in \lama{} for the different context strategies. These results are shown in Table \ref{tab:table_nsp}. We see that for \ret{}, \gen{} and \ora{}, \nsp{} classifications are high, suggesting \berttext{} finds the segments to be contiguous, and hence useful to condition upon. However, for \adv{}, very few $(\context,\question)$ pairs are classified as ``next sentences", suggesting \berttext{} may condition on them less. Additional evidence for our \nsp{} adversarial robustness hypothesis is given in Figure \ref{fig:nsp_correlation}. 
Here we compute the absolute difference in probability that \berttext{} places on the correct answer upon including context $\left|| P_{LM}(a | q) - P_{LM}(a | q + c) \right||$, and plot it against \nsp{} probability $P_{NSP}(q, c)$. We see that for  adversarial, retrieved and generated contexts, increasing \nsp{} probability is associated with greater change in true answer probability upon including context.\footnote{Each context method has different \nsp{} statistics, (\eg{} the generated contexts have very high \nsp{} probabilities on average) but the trend is consistent---higher NSP scores co-occur with greater changes in correct answer probability  }

\subsection{Generated context}

By generating context from a LM we aim at assessing the performance of a solution purely based on knowledge implicitly stored in the parameters of the underlying neural networks. Although the overall results of \gen{} are lower than the context-free baseline, some interesting insights emerge from our analysis. First, generated context improves performance for 13 relations and overall for $7\%$ of the questions on T-REx (Table \ref{tab:origintrex}). This demonstrates that autoregressive language models can generate relevant context and potentially serve as unsupervised IR systems. They do, of course, generate also irrelevant or factually wrong information. What is interesting is that BERT associates high \nsp{} probabilities with generated contexts---for BERT, the generation is always a plausible continuation of the question. This inhibits the selective behaviour of BERT with respect to the context, and hurts performance when the generation is noisy, irrelevant or wrong. 
%We hypothesize that the NSP scores are high for generated contexts exactly because they were generated by another language model which believed these tokens to be the most likely ones to follow the query statement. Unfortunately, the tokens often do not contain any useful information (and in fact, they may be misleading).

Table \ref{tab:trexexamplesbert} shows three examples for the generation of BERT-large for adversarial,  generated, retrieved and oracle context-enriched questions.

\newcolumntype{b}{X}
\newcolumntype{s}{>{\hsize=.22\hsize}Xl}

\begin{table}[]
\renewcommand{\arraystretch}{1.3}
\centering
\small
\resizebox{\textwidth}{!}{   
\begin{tabularx}{\linewidth}{lb p{2.8cm}}
\toprule
& Query & Predictions \\

\midrule

\multicolumn{2}{l}{ [P101] \textsc{Allan Sandage works in the field of \underline{\hspace{2em}} .}}
&
\makecell[tl]{
engineering {\scriptsize [-3.1]} \\
} \\

\belowbaseline[0ex]{\rotatebox[origin=c]{90}{{\scriptsize \textsc{adv}}}}
&
$q$ [SEP]  According to Gould, classical Darwinism encompasses three essential core commitments: Agency, the unit of selection, which for Charles Darwin was the organism, upon which natural selection ... \hfill  {\scriptsize [0.0]}
&
\makecell[tl]{
psychology {\scriptsize [-2.8]} \\
economics {\scriptsize [-3.4]} \\
anthropology {\scriptsize [-3.5]} \\
} \\

\belowbaseline[0ex]{\rotatebox[origin=c]{90}{{\scriptsize \textsc{gen}}}}
&
$q$ [SEP]   How many hours a week does he work? Does he get paid? How much does he get paid? How much does he get paid? He does not have a car. \hfill  {\scriptsize  [1.0]}
&
\makecell[tl]{
finance {\scriptsize [-2.1]} \\
engineering {\scriptsize [-3.4]} \\
advertising {\scriptsize [-3.4]} \\
} \\

\belowbaseline[0ex]{\rotatebox[origin=c]{90}{{\scriptsize \textsc{ret}}}}
&
$q$ [SEP]  In 1922 John Charles Duncan published the first three variable stars ever detected in an external galaxy, variables 1, 2, and 3, in the Triangulum Galaxy (M33). These were followed up by Edwin ... \hfill  {\scriptsize [1.0]}
&
\makecell[tl]{
\textbf{astronomy {\scriptsize [-0.0]} } \\
physics {\scriptsize [-5.5]} \\
observation {\scriptsize [-7.3]} \\
} \\

\belowbaseline[0ex]{\rotatebox[origin=c]{90}{{\scriptsize \textsc{ora}}}}
&
$q$ [SEP]  He currently works at the Institute of Astronomy in Cambridge; he was the Institute's first director.Educated at the University of Cambridge, in 1962 he published research with Olin Eggen and Allan ... \hfill  {\scriptsize [1.0]}
&
\makecell[tl]{
\textbf{astronomy {\scriptsize [-0.0]} } \\
physics {\scriptsize [-4.0]} \\
galaxies {\scriptsize [-5.5]} \\
} \\

\midrule

\multicolumn{2}{l}{ [P279] \textsc{Interleukin 6 is a subclass of \underline{\hspace{2em}} .}}
&
\makecell[tl]{
proteins {\scriptsize [-0.2]} \\
} \\

\belowbaseline[0ex]{\rotatebox[origin=c]{90}{{\scriptsize \textsc{adv}}}}
&
$q$ \sep  First built in 1893 by Chinese residents of Nagasaki with the support of the Qing Dynasty government, the shrine was designed to serve as a place of worship and learning for the Chinese ... \hfill  {\scriptsize [0.0]}
&
\makecell[tl]{
proteins {\scriptsize [-0.2]} \\
\textbf{protein {\scriptsize [-3.1]} } \\
DNA {\scriptsize [-3.7]} \\
} \\

\belowbaseline[0ex]{\rotatebox[origin=c]{90}{{\scriptsize \textsc{gen}}}}
&
$q$ \sep   Okay, let's get this out of the way. The Interleukin 6 (IL-6) is an interleukin-6 receptor (IL-6R) that plays a key role in the immune system. Intra-leukin-6 (IL-6R) is an interleukin-6 receptor (IL ... \hfill  {\scriptsize [1.0]}
&
\makecell[tl]{
proteins {\scriptsize [-0.6]} \\
receptors {\scriptsize [-1.6]} \\
antibodies {\scriptsize [-2.2]} \\
} \\

\belowbaseline[0ex]{\rotatebox[origin=c]{90}{{\scriptsize \textsc{ret}}}}
&
$q$ \sep  In particular, the increase in levels of IL-6 (interleukin 6), a myokine, can reach up to one hundred times that of resting levels. Depending on volume, intensity, and other training factors, the IL ... \hfill  {\scriptsize  [1.0]}
&
\makecell[tl]{
insulin {\scriptsize [-1.9]} \\
IL {\scriptsize [-2.1]} \\
proteins {\scriptsize [-2.4]} \\
} \\

\belowbaseline[0ex]{\rotatebox[origin=c]{90}{{\scriptsize \textsc{ora}}}}
&
$q$ \sep  It is a cardiac hypertrophic factor of 21.5 kDa and a protein member of the IL-6 cytokine family. This protein heterodimerizes with interleukin 6 signal transducer to form the type II oncostatin M ... \hfill  {\scriptsize [1.0]}
&
\makecell[tl]{
proteins {\scriptsize [-0.7]} \\
\textbf{protein {\scriptsize [-1.5]} } \\
insulin {\scriptsize [-2.4]} \\
} \\

\midrule

\multicolumn{2}{l}{ [P413] \textsc{Giacomo Tedesco plays in \underline{\hspace{2em}} position .}}
&
\makecell[tl]{
center {\scriptsize [-2.2]} \\
} \\

\belowbaseline[0ex]{\rotatebox[origin=c]{90}{{\scriptsize \textsc{adv}}}}
&
$q$ \sep  On July 31, 2009 he was traded from the Tigers to the Seattle Mariners along with fellow pitcher Luke French for veteran pitcher Jarrod Washburn. On July 31, 2009 he was traded from the Tigers to ... \hfill  {\scriptsize  [0.03]}
&
\makecell[tl]{
center {\scriptsize [-1.5]} \\
centre {\scriptsize [-2.4]} \\
forward {\scriptsize [-2.6]} \\
} \\

\belowbaseline[0ex]{\rotatebox[origin=c]{90}{{\scriptsize \textsc{gen}}}}
&
$q$ \sep   How much does he play? He can play fullback, wing or centre. He can also play on the wing. Tedesco can also play in the halves. Tedesco can play in the halves. \hfill  {\scriptsize [1.0]}
&
\makecell[tl]{
fullback {\scriptsize [-1.4]} \\
centre {\scriptsize [-2.1]} \\
wing {\scriptsize [-3.6]} \\
} \\

\belowbaseline[0ex]{\rotatebox[origin=c]{90}{{\scriptsize \textsc{ret}}}}
&
$q$ \sep  Giovanni Tedesco has two brothers who are also football players, Salvatore (formerly of Perugia and Lucchese) and Giacomo, who is playing for Reggina. \hfill  {\scriptsize [1.0]}
&
\makecell[tl]{
\textbf{midfielder {\scriptsize [-1.2]} } \\
forward {\scriptsize [-1.8]} \\
midfield {\scriptsize [-2.3]} \\
} \\

\belowbaseline[0ex]{\rotatebox[origin=c]{90}{{\scriptsize \textsc{ora}}}}
&
$q$ \sep  Giacomo Tedesco (born February 1, 1976 in Palermo) is a former Italian football (soccer) midfielder. Giacomo Tedesco (born February 1, 1976 in Palermo) is a former Italian football (soccer) midfielder ... \hfill  {\scriptsize  [1.0]}
&
\makecell[tl]{
\textbf{midfielder {\scriptsize [-0.7]} } \\
forward {\scriptsize [-2.2]} \\
defender {\scriptsize [-2.4]} \\
} \\

\bottomrule 
\end{tabularx}
}
\caption{Examples of generation for BERT-large. We report the top three tokens predicted with the associated log probability (in square brackets) for adversarial (\textsc{adv}), generated (\textsc{gen}), retrieved (\textsc{ret}) and oracle (\textsc{ora}) context-enriched questions. NSP probability (in square brackets) reported at the end of each statement. }
\label{tab:trexexamplesbert}
\end{table}

\section{Discussion}
In this section we discuss some of our findings and their implications.

\paragraph{Re-examining NSP} The Next Sentence Prediction task has been extensively explored \cite{devlin2018bert,liu2019roberta,yang2019xlnet,Lan2019ALBERTAL} with the apparent consensus that it is not helpful for downstream fine-tuning accuracy. 
Our findings, in contrast, suggest that it is important for robust exploitation of retrieved context for unsupervised tasks.
Basing design decisions with a limited set of downstream uses when designing general purpose pre-trained models may well us lead to less flexible models. As a community, we should continue to strive for greater diversity in our criteria and possible use-cases for assessing such models \cite{DBLP:journals/corr/abs-1912-13283}.

\paragraph{Practical Takeaways} Section \ref{sec:results} shows that \berttext{} has a very different behaviour when inputs are processed with one or two segments. Practitioners should thus ensure that they thoroughly ablate segmentation options. The consistent improvement upon including retrieved context also suggests that it may be possible to get performance boosts in many other tasks by the trivial incorporation of retrieved documents, even when such documents are not strictly required for the task. We leave investigating this for future work. 

\paragraph{Comparison with \drqa{}} We demonstrate that \berttext{} with retrieved context and no fine-tuning performs on par with \drqa{} on the \lama{} probe, but it is worth discussing this comparison further. Firstly, it is encouraging that an unsupervised system performs just as well as a system that requires significant supervision such as \drqa{}. We further note that LMs are \emph{abstractive} models, whereas \drqa{} is \emph{extractive}, confined to returning answers that are spans of retrieved context. However, it is worth stating that \lama{} only requires single token answers. Generating an arbitrarly long sequence of contiguous tokens from bidirectional LMs like \berttext{} and \robertatext{} is not trivial, but extractive QA models handle such cases by considering spans of text of varying lengths. Finally, while we have chosen \drqa{} as our baseline to compare to recent work, there exist several more sophisticated supervised open-domain QA models that outperform it on a variety of open-domain QA tasks \cite{lee-etal-2019-latent,yang-etal-2019-end-end,realm}.

\paragraph{Unsupervised Question Answering} Our work is part of growing body of work that demonstrate that unsupervised question answering is not only possible, but beginning to reach and even outperform some standard supervised baselines. \citet{radford2019language} and \citet{lewis-etal-2019-unsupervised} demonstrate non-trivial performance on CoQA~\cite{reddy-etal-2019-coqa} and SQuAD~\cite{rajpurkar2016squad} respectively, and \cite{yadav-etal-2019-alignment} achieve SoTA results using an unsupervised method for multi-choice QA on ARC~\cite{DBLP:journals/corr/abs-1803-05457}. Taken together, these recent findings suggest that powerful and flexible unsupervised QA systems could soon be a reality, bringing with them many advantages including avoiding biases that often plague smaller datasets by incorporating knowledge from much larger corpora and greater abilities to combine and abstract pieces of information from different sources.

\section{Conclusion}

We demonstrated a simple technique to greatly improve factual unsupervised cloze QA by providing context documents as additional inputs. We used oracle documents to establish an upper bound to this improvement, and found that using off-the-shelf information retrieval is sufficient to achieve performance on par with the supervised \drqa{} system. We also investigated how brittle language models' factual predictions were to noisy and irrelevant context documents, and found that \berttext{}, when featurized appropriately, is very robust. We provide evidence that this robustness stems from the Next Sentence Prediction pre-training task. 

\bibliography{LAMA}
\bibliographystyle{plainnat}

\end{document}